\documentclass[journal,twoside,web]{ieeecolor}
\usepackage{generic}

\usepackage{cite}
\usepackage{amsmath,amssymb,amsfonts}
\usepackage{algorithmic}
\usepackage{graphicx}
\usepackage{textcomp}
\usepackage{mathtools}

\usepackage{hyperref}

\usepackage{booktabs}
\usepackage{tablefootnote}

\usepackage{dblfloatfix}    
\usepackage{siunitx}

\usepackage{bm}
\usepackage{siunitx}

\def\RH{R_\mathrm{H}}
\def\RL{R_\mathrm{L}}
\def\IH{I_\mathrm{H}}
\def\IL{I_\mathrm{L}}
\def\US{V_\mathrm{S}}
\def\UR{V_\mathrm{R}}

\def\RHn{R_{\mathrm{H,}n}}
\def\RHnp{R_{\mathrm{H,}n+1}}
\def\RLn{R_{\mathrm{L,}n}}
\def\USn{V_{\mathrm{S,}n}}
\def\URn{V_{\mathrm{R,}n}}

\def\mrh{\mu({R_{H,n}})}
\def\mus{\mu({V_{S,n}})}
\def\mrl{\mu({R_{L,n}})}
\def\mur{\mu({V_{R,n}})}
\def\srh{\sigma({R_{H,n}})}
\def\sus{\sigma({V_{S,n}})}
\def\srl{\sigma({R_{L,n}})}
\def\sur{\sigma({V_{R,n}})}

\def\G{\bm{\Gamma}}

\def\IH{I_{\text{H}}}
\def\IL{I_{\text{L}}}

\def\IHRSnp1{I_{\text{HRS,n+1}}}
\def\ILRSnp1{I_{\text{LRS,n+1}}}
\def\Umax{V_{\text{max}}}

\newcommand{\inv}{^{\raisebox{.2ex}{$\scriptscriptstyle\text{-}1$}}}

\usepackage{xspace}
\def\A{\textbf{(A)}\xspace}
\def\B{\textbf{(B)}\xspace}
\def\C{\textbf{(C)}\xspace}
\def\D{\textbf{(D)}\xspace}
\def\E{\textbf{(E)}\xspace}
\def\F{\textbf{(F)}\xspace}
\def\Gee{\textbf{(G)}\xspace}

\def\BibTeX{{\rm B\kern-.05em{\sc i\kern-.025em b}\kern-.08em
    T\kern-.1667em\lower.7ex\hbox{E}\kern-.125emX}}
\begin{document}
\title{Synaptogen: A cross-domain generative device model for large-scale neuromorphic circuit design}
\author{Tyler Hennen, Leon Brackmann, Tobias Ziegler, Sebastian Siegel, Stephan Menzel, Rainer Waser, \hbox{Dirk J. Wouters}, and Daniel Bedau
\thanks{
This work was supported by the German Federal Ministry of Education and Research through the projects NEUROTEC~II (16ME0399, 16ME0398K) and NeuroSys (03ZU1106AA, 03ZU1106AB).}
\thanks{T. Hennen  (e-mail: t.hennen@iwe.rwth-aachen.de), L. Brackmann, T. Ziegler, and D.J. Wouters are with the Institut für Werkstoffe der Elektrotechnik 2 (IWE2), RWTH Aachen University, Sommerfeldstraße 18/24 52074 Aachen, Germany.}
\thanks{S. Siegel and S. Menzel are with the Peter Grünberg Institute (\hbox{PGI-7}), Forschungszentrum Jülich, Wilhelm-Johnen-Straße
52428 Jülich, Germany.}
\thanks{R. Waser is with both IWE2 and PGI-7.}
\thanks{D. Bedau (email: daniel.bedau@wdc.com) is with Western Digital Corporation, 5601 Great Oaks Parkway, San Jose, CA 95119.}}

\maketitle

\begin{abstract}
We present a fast generative modeling approach for resistive memories that reproduces the complex statistical properties of real-world devices.
To enable efficient modeling of
analog circuits, the model is
implemented in
Verilog-A.
By training on extensive measurement data of integrated 1T1R arrays (6,000 cycles of 512 devices),
an autoregressive stochastic process accurately accounts for the cross-correlations between
the switching parameters, while non-linear transformations ensure agreement with
both cycle-to-cycle (C2C) and \mbox{device-to-device} (D2D) variability.
Benchmarks show that this statistically comprehensive model achieves read/write throughputs exceeding those of
even highly simplified and deterministic compact models.

\end{abstract}


\section{Introduction}

A pressing challenge for large-scale simulations of neuromorphic systems is the availability of suitable synaptic device models for resistive memories such as ReRAM~\cite{b0}. For
applications, it is important
to capture
the complex stochastic behavior of the devices, and
models need to be fast enough to simulate millions 
of cells at once
to handle modern neural network
circuits.
To this end, computationally lightweight generative models can be trained on electrical characteristics of fabricated devices,
providing high speed simulations of large networks with unprecedented statistical accuracy~\cite{b1}.

While our previous work
focused on large-scale simulations in
high-level programming languages, here we present a circuit-level model
implemented in the hardware description language Verilog-A, which is necessary
to bridge the divide between the machine learning (ML) and analog circuit simulation
domains.
The model was expanded to cover a device configuration with access transistors (1T1R),
and we introduce a measurement protocol for collecting the necessary
training data on integrated memory arrays.
The stochastic modeling approach
closely
captures the distributions, cross-correlations, and history dependence
of ReRAM switching parameters as the devices are cycled (C2C),
and has an extended treatment of
how those statistics vary
between the different devices on the chip (D2D).
The
resulting
device
model is far more statistically comprehensive than existing compact models and significantly outperforms them in read/write benchmarks for both independent devices and for crossbar arrays.
In circuit simulations, we demonstrate
weight programming and readout of crossbars with up to 256$\times$256 and 1024$\times$1024 devices respectively, the feasibility of which has not been shown previously.

\section{Methods}

\subsection{Electrical measurements}

\begin{figure}
\centering
\includegraphics[width=0.7\linewidth]{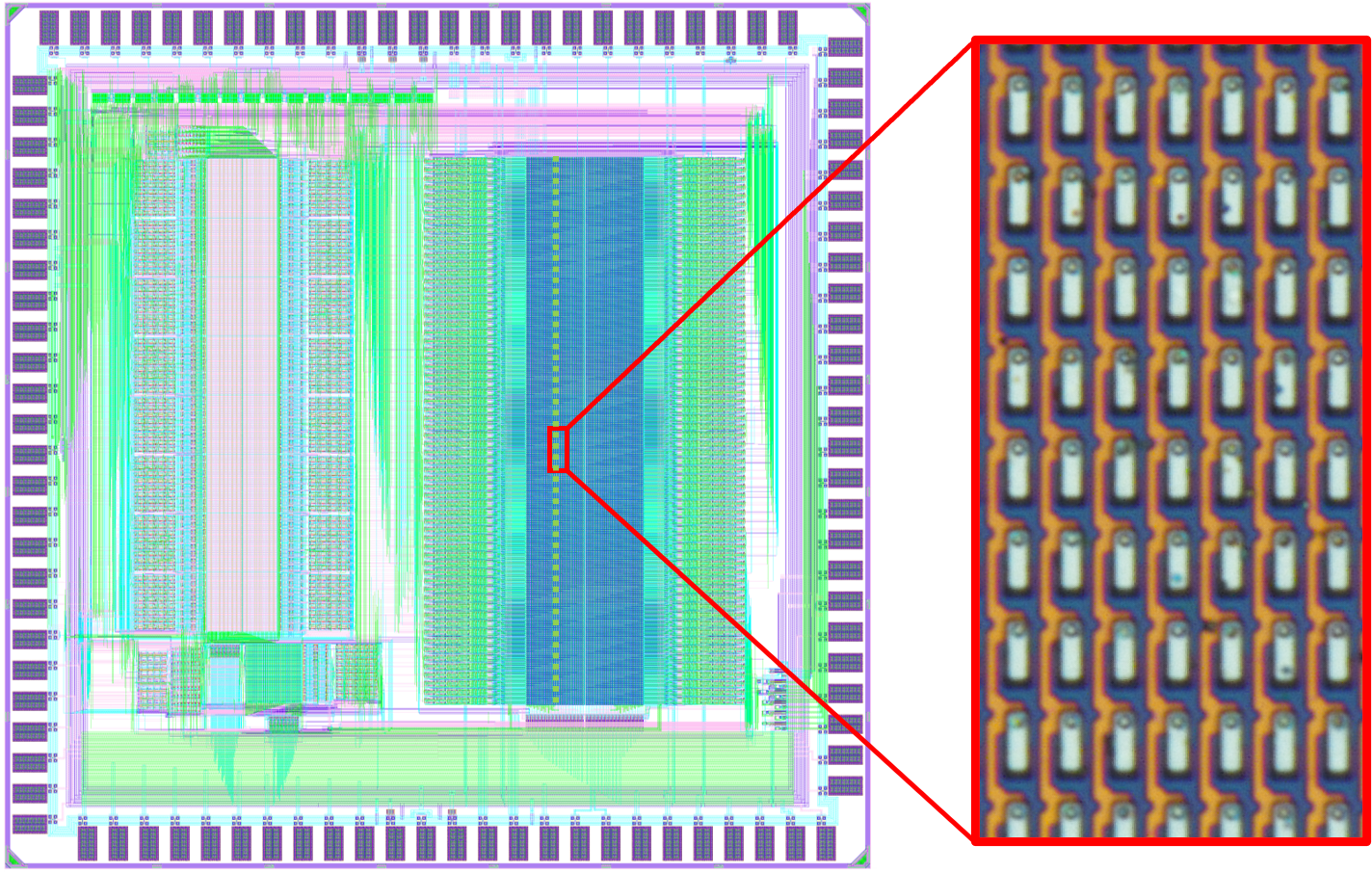}
\caption{The ReRAM chip layout in the MAD200 process design kit (left)
and an optical image of the fabricated 1T1R ReRAM array (right).}\label{fig:1T1R_array_optical}
\end{figure}

\begin{figure}
\centering
\includegraphics[width=0.8\linewidth]{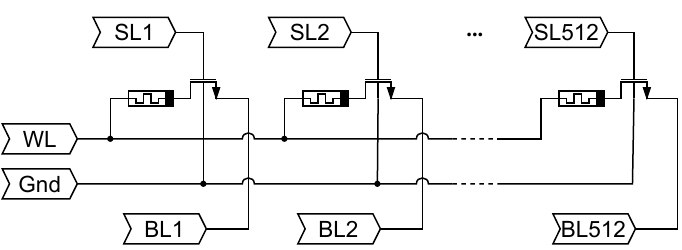}
\caption{Simplified circuit diagram showing a connected vector of 512 1T1R ReRAM
devices which were individually selected for measurement of model training data.
Transistor bodies (``Gnd'' terminal) were biased to -1.8 V as bipolar voltage sweeps were applied
to the WL and current was measured at the respective BL (at 0~V).}\label{fig:1T1R_array_schematic}
\end{figure}

An integrated ReRAM chip was obtained through the manufacturing broker Circuits Multi-Projects (CMP) and
used for electrical measurements.
A 512$\times$32 1T1R crossbar array was part of a custom layout within the Memory Advanced Demonstrator 200mm (MAD200) design environment~(Fig.~\ref{fig:1T1R_array_optical}).
Select logic and access transistors were implemented in the HCMOS9A
STMicroelectronics 130~nm CMOS process, and ReRAM devices with material
stack TiN/HfO$_2$/Ti were deposited in a post-process by CEA-LETI~\cite{b2}.
Each ReRAM device in the array is connected in series with an integrated
common-source N-channel MOSFET in a standard 1T1R configuration.
The 512 bit lines and corresponding select lines (SLs)
are each internally multiplexed to single output pins, 
whereas the 32 word lines (WLs) are directly routed to individual
pins.
The packaged chip was mounted on a custom printed circuit board (PCB)
providing a PC interface via the digital outputs of a USB data acquisition board
whereby devices can be individually addressed for measurement.
In this work, a total of 512 devices sharing a single WL in the array were
sequentially selected 
to collect training data~(Fig.~\ref{fig:1T1R_array_schematic}).

High speed measurements were performed using external generating and sampling
equipment connected to the PCB over 50~$\unit{\ohm}$ lines.
In order to collect bipolar switching cycles continuously with a single driving signal, an unusual 1T1R biasing was necessary. 
The chip substrate (and FET body) was biased to -1.8~V relative to signal ground
and the gate was biased to 1.35~V while a bipolar driving signal was applied
to the WL and current was measured at the BL through a 50~$\unit{\ohm}$ shunt to
0~V.
Devices were formed by a single \mbox{3~V} amplitude 1~ms triangle pulse
before being cycled by a continuous triangle waveform between
-1.5~V and 2~V with 1~ms period. Preconditioning cycles were initially applied
to each cell before collecting 6,000 switching current vs.\ voltage ($I, V$)
traces for each of the 512 devices.

\subsection{Statistical modeling}

The core concept of the generative device model is to first extract important features (i.e.\ resistance
and voltage threshold levels) from each cycle of the training data, then learn to
efficiently generate new samples with very similar statistical properties.  Using the
generated features as a guide,
we approximate the
$I(V)$
dependence for simulated cells according to the
voltage sequence applied to them.

\subsubsection{Feature generation}

The chosen features to model are extracted from the raw data
and organized into
vector time series
\begin{equation}
\bm{x}_{n,m} = \begin{bmatrix}\RH \\ \US \\ \RL \\ \UR\end{bmatrix}_{n,m} 
\end{equation}
for each cycle number $n \in [1, N]$
and
device number \mbox{$m \in [1, M]$}.
The feature vectors are arranged from top to bottom in the order that they occur in the measurement;
$\RH$ is the resistance of the high resistance state (HRS), $\US$ is the
voltage of the SET transition, $\RL$ is the resistance of the low resistance
state (LRS), and $\UR$ is the voltage at the start of the RESET transition.
The details of this feature extraction are documented in~\cite{b1}.

\begin{figure}[t]
\centering
\includegraphics[width=0.7\linewidth]{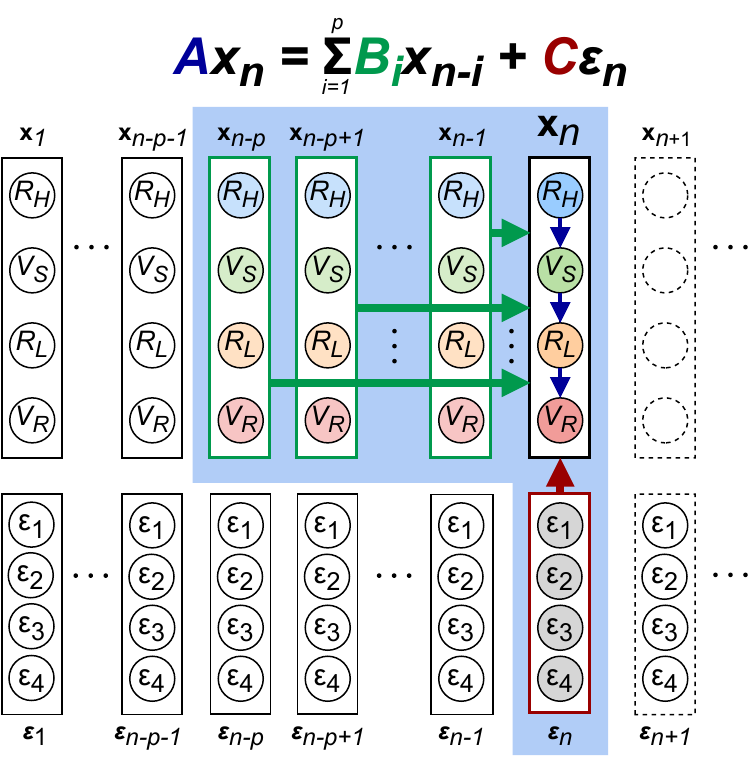}
\caption{
Graphical depiction of the VAR($p$)
base process used to reproduce memory
cycling statistics. Past features within cycle
range $p$ have a linear deterministic impact on
future values, and a 4-dimensional white noise
process $\epsilon_n$ contributes stochasticity to each feature.
    }\label{fig:var}
\end{figure}

Feature vector generation is based on a discrete vector autoregressive (VAR)
stochastic process (Fig.~\ref{fig:var}), which captures the cycle
history dependence and the correlations between features~\cite{b6}.
A VAR($p$) models the $n$th feature vector as a linear function of
past values
within cycle range $p$ and is driven by 
4-dimensional white noise $\bm{\epsilon}_n$.
The stochastic process has the easily computable form,
\begin{equation}
\bm{A} \bm{x}_n = \sum_{i=1}^p \bm{B}_i \bm{x}_{n-i} + \bm{C} \bm{\epsilon}_n,
\end{equation}
where $\bm{A}$, $\bm{B}_i$, and $\bm{C}$ are $4 \times 4$ weight matrices subject to training.

\begin{figure*}[ht!]
\begin{center}
  \includegraphics[width=\linewidth]{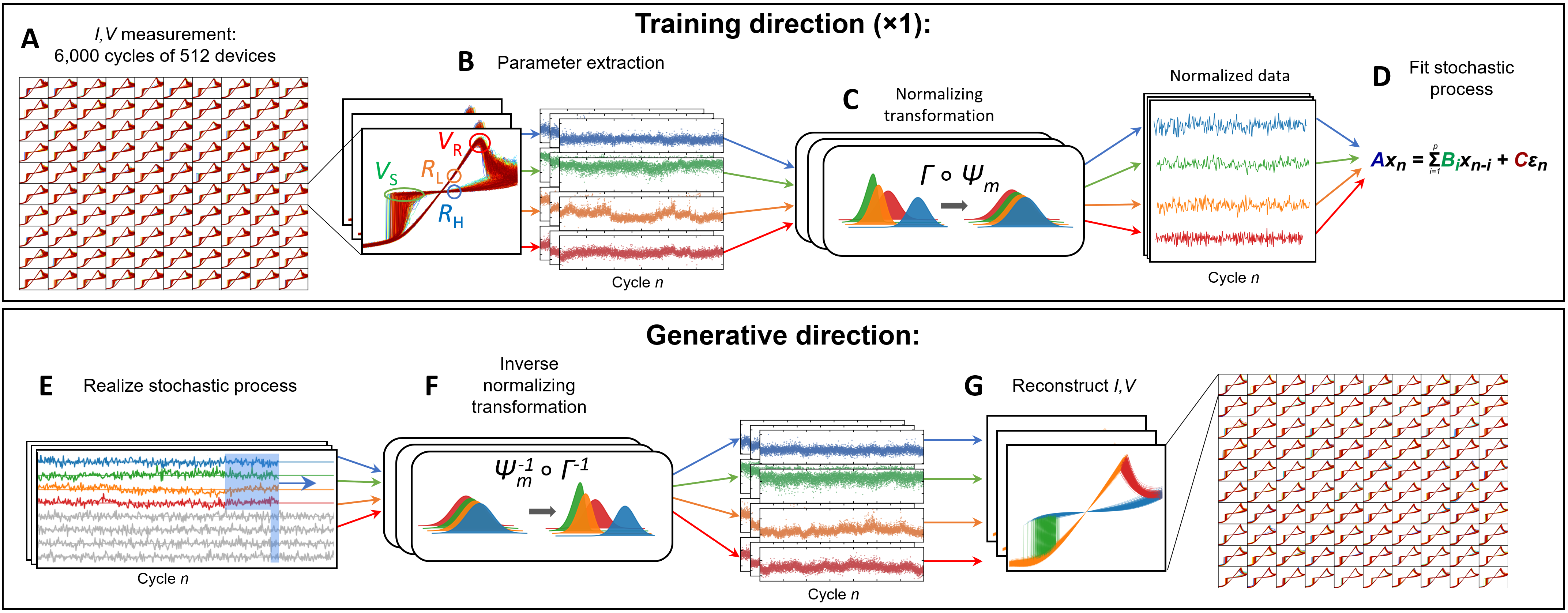}
\end{center}
\caption{
An overview of the generative modeling approach. Training direction: \A collect
$I, V$ data ($N$ cycles $\times$ $M$ devices), \B extract feature vectors, \C learn a
distribution of normalizing transformations, \D fit a stochastic process (VAR)
to the normalized data.  Generative direction: \E realize an independent VAR process for each simulated cell,
\F apply device-specific random de-normalizing transformations to the VAR outputs, \Gee as voltages are applied, reconstruct $I, V$ dependence of each cell.
}\label{fig:generative_model}
\end{figure*}

\begin{figure}
  \centering
  \includegraphics[]{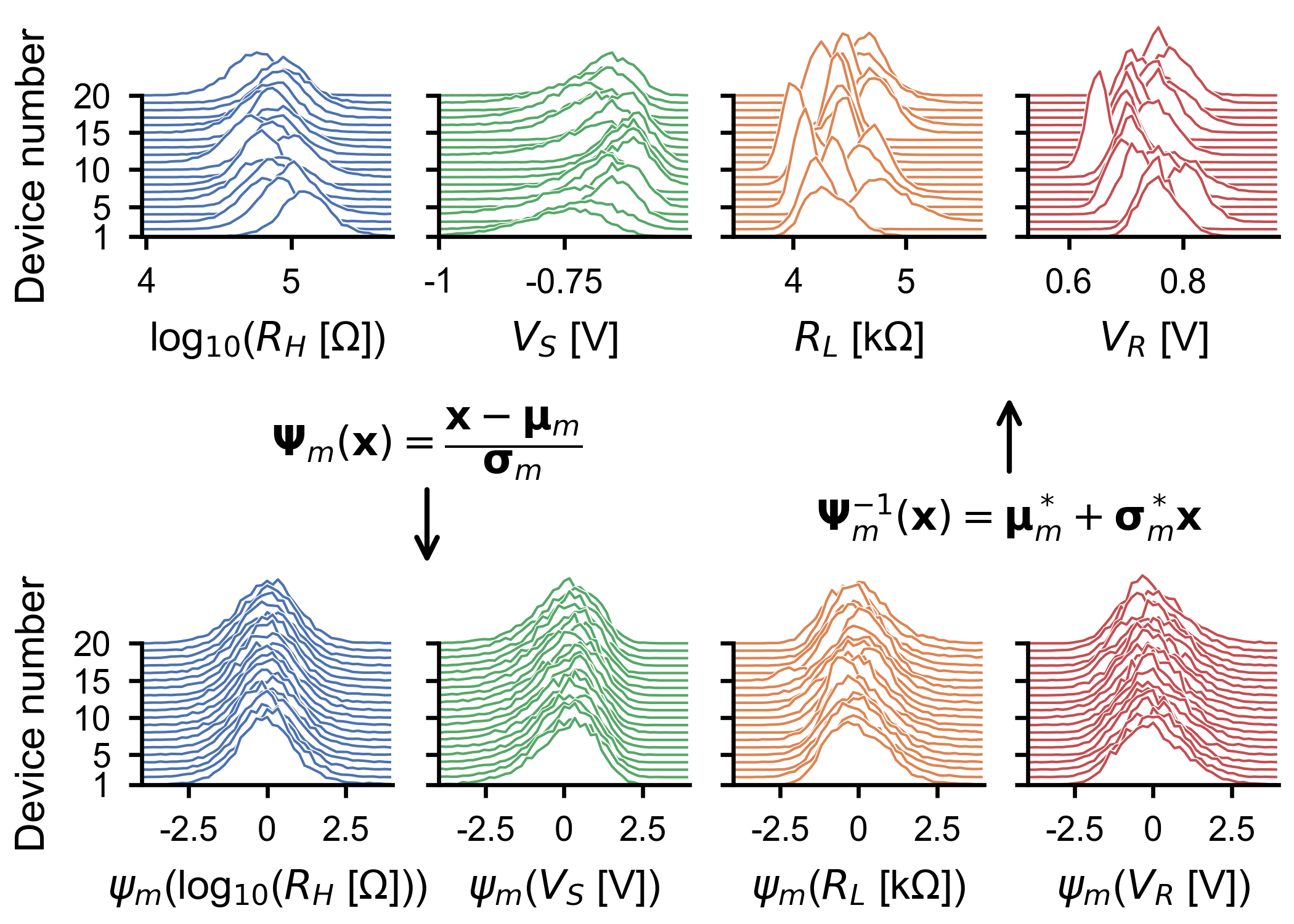}
  \caption{
A standardizing affine transformation applied to a representative sample of 20 different devices.
The forward transformation $\bm{\Psi}_m$ is applied in the
training direction as a first step to normalize the feature distributions.
Here, $\bm{\mu}_m$ and $\bm{\sigma}_m$ are the sample 
means and standard deviations of the feature vectors for device $m$ across all cycles.
The inverse transformation is used in the
generative direction, where $\bm{\mu}_m^*$ and $\bm{\sigma}_m^*$ are sampled from a
distribution estimated from the entire training set.}\label{fig:affine}
\end{figure}

To map the normally distributed output of the VAR process to the joint
empirical distribution
measured across cycles and across devices,
we 
apply a sequence of invertible transformations.
The parameters of these transformations are learned in a single
training pass in which the generative process is carried out in
reverse~(Fig.~\ref{fig:generative_model}). 
Thereby, the marginal distributions of the
extracted features are
normalized in two steps.
First, the device-specific mean and variance over the sampled cycles are
standardized using an affine transformation $\bm{\Psi}_m$~(Fig.~\ref{fig:affine}).
Then, to
further shape
the intermediate
probability densities into normal distributions, the affine transformation is
followed by a parameterized, non-linear
quantile transform
$\bm{\Gamma}$.
A VAR($p$) process is then fit to the normalized data using least squares regression.

\begin{figure}
  \centering
  \includegraphics[]{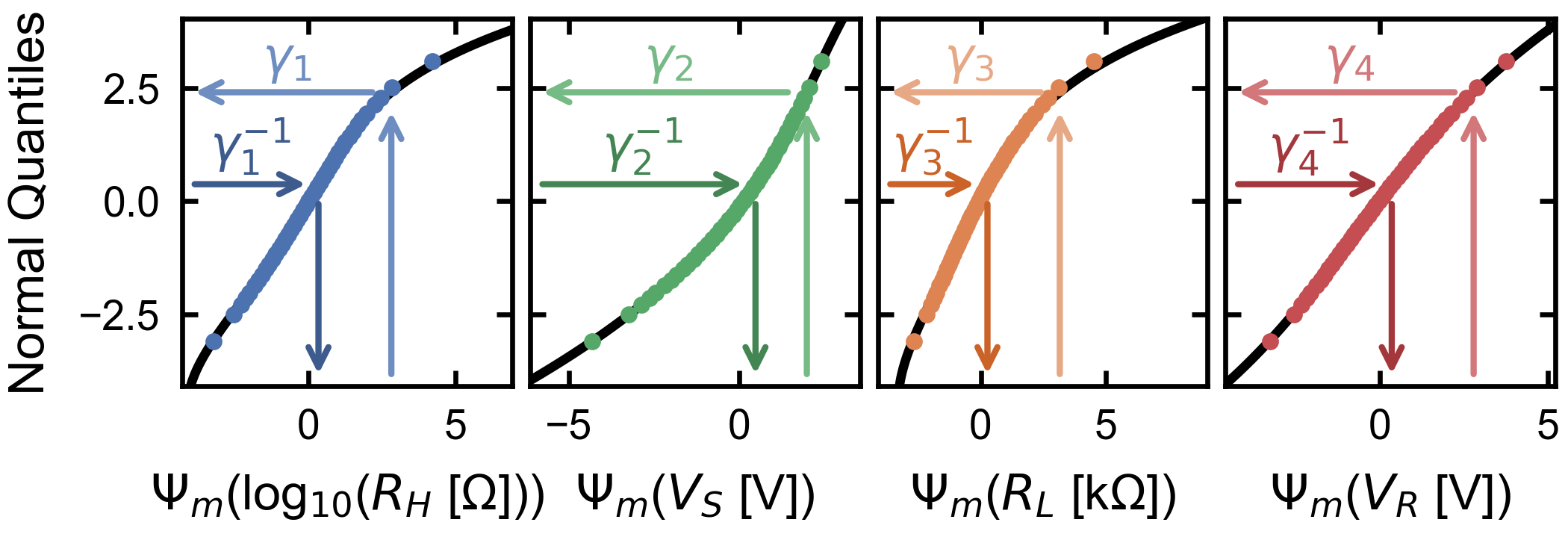}
  \caption{The elementwise non-linear quantile transform $\bm{\Gamma}$ adapted to the training data. The
inverse transformations are polynomial functions $\gamma_i$ designed for fast evaluation during
the generative process. The non-linearity allows the model to reproduce the non-normal,
asymmetric distributions presented by the training data.}\label{fig:gamma}
\end{figure}

In the generative direction, the learned transformations are inverted and applied to
independent realizations of the
VAR process for each simulated device.
The
normalizing map
$\bm{\Gamma}$ is
defined
such that its inverse
consists of element-wise polynomial evaluations,
\begin{equation}
\bm{\G}\inv({\bm{x}})
=
\begin{bmatrix*}[l]
\gamma_1({R}_{H})\\ 
\gamma_2({V}_{S})\\ 
\gamma_3({R}_{L})\\ 
\gamma_4({V}_{R})
\end{bmatrix*}
\end{equation}
where $\gamma_i$ are 4th degree polynomials and are visualized in Fig.~\ref{fig:gamma}.

\begin{figure*}
\centering
  \includegraphics[]{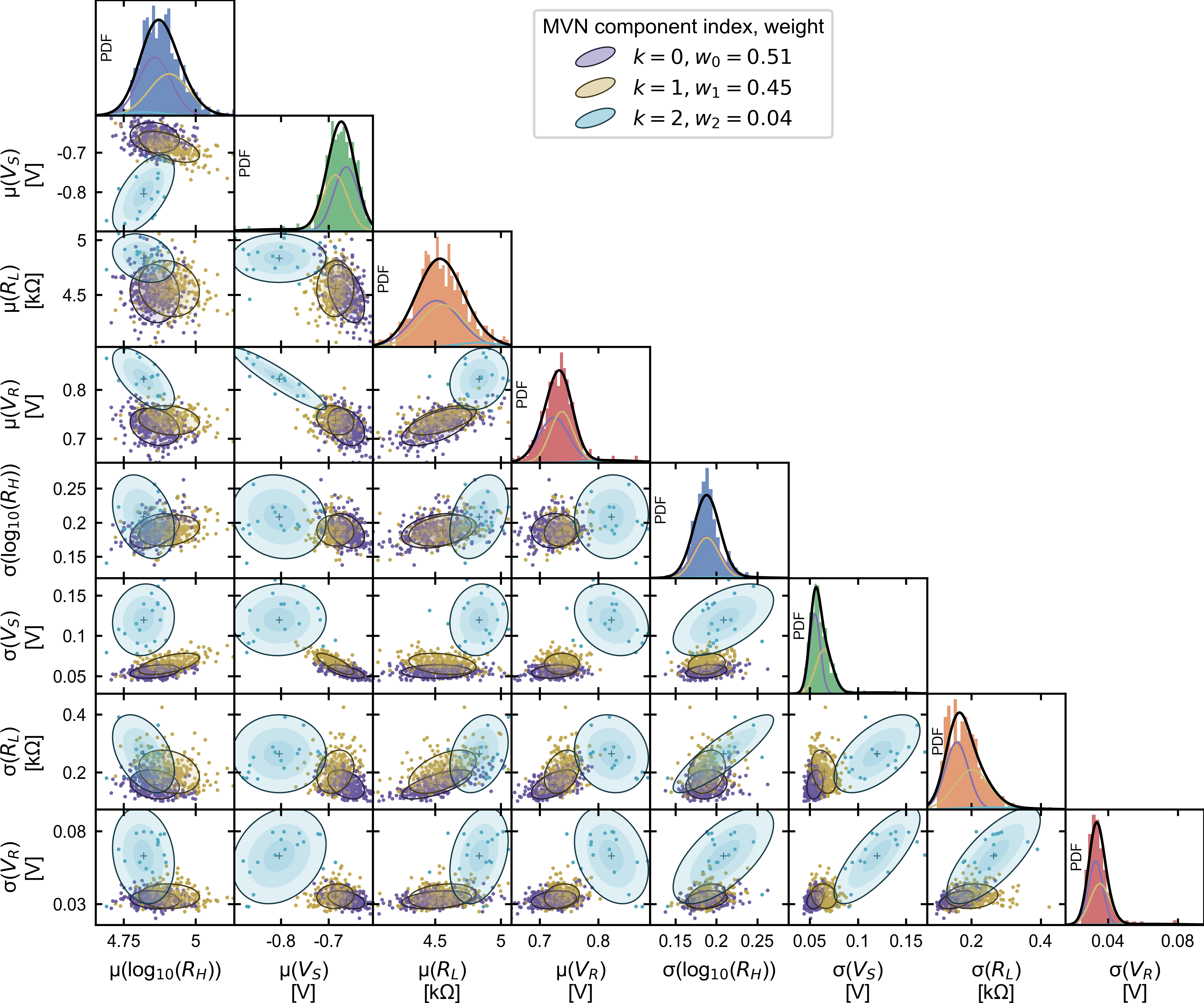}
  \caption{Correlative scatterplot of the
  feature
  means ($\mu$) and standard deviations ($\sigma$) over all cycles
  of devices in the training set.
  The 512 datapoints are fit
and classified by a GMM with three Gaussian components (purple, yellow, and teal), allowing sampling of new
vectors for the generative model. Component $k=2$ (teal) captures the multivariate structure of a device defect occurring in 4\%
of devices.  The diagonal subplots show that the weighted addition of the marginal probability distribution functions (PDFs) of the three components
closely fit the histograms of the input data.
}\label{fig:DtD_distributions}
\end{figure*}

To
restore
device-specific offsets and scales to the generated features, we invert $\bm{\Psi}_m$
by approximating the distribution of an 8-dimensional block vector of 
sampled C2C means ($\mu$) and standard deviations ($\sigma$),
\begin{equation}
\bm{S}_m
=
\left[\begin{array}{cc}
\bm{\mu}_m  \\
\bm{\sigma}_m \\
\end{array}\right]
=
\begin{bmatrix*}[l] \mrh \\ \mus \\ \mrl \\ \mur \\ \srh \\ \sus \\ \srl \\ \sur \end{bmatrix*}_m.
\end{equation}
This distribution is represented by a superposition of multivariate normal (MVN) distributions, which is known as
a Gaussian mixture model (GMM).
A GMM is cheap to sample from and allows a close fit of the covariance structure of the main cluster of $\bm{S}_m$ datapoints.
The GMM also captures the structure of statistical abnormalities that occur (i.e.\ defective devices),
which may have a disproportionate
impact on system performance.
A three-component GMM, denoted
\begin{equation}
  \bm{S}^*_m
=
\left[\begin{array}{cc}
\bm{\mu}^*_m  \\
\bm{\sigma}^*_m \\
\end{array}\right],
\end{equation}
is fit to the empirical distribution by the expectation–maximization algorithm using k-means
initialization and is visualized in Fig.~\ref{fig:DtD_distributions}.

\subsubsection{Modeling the $I(V)$ dependence}

The non-linear $I(V)$ state for each cell is modeled as a linear combination of
two static, global limiting polynomials $I_\mathrm{H}(V)$ and $I_\mathrm{L}(V)$
(degree 5 and 6 respectively),
whose coefficients are estimated from the training data.
This way, the model can reproduce a wide variety of asymmetric non-linearities
in both high and low resistance states, and can
also absorb the non-linearity
of series transistors when trained on 1T1R data.

Resistance levels of the devices 
are tracked by continuous state variables $r_m \in (0, 1)$,
which represent the degree of mixing between the pre-defined limiting polynomials.
The current as a function of voltage for device $m$
assumes the form
\begin{equation}
I_m(r_m, V) = r_m I_\mathrm{H}(V) + (1-r_m) I_\mathrm{L}(V).
\end{equation}
The state variable corresponding to each generated resistance level $R$
is calculated using the function
\begin{equation}
r(R) = \frac{\IL(V_0) - V_0 R\inv} {\IL(V_0) - \IH(V_0)},
\end{equation}
which uniquely sets the static resistance of the device (evaluated at $V_0 = 0.2$~V) equal to $R$.

\begin{figure}[h!]
\centering
\includegraphics[]{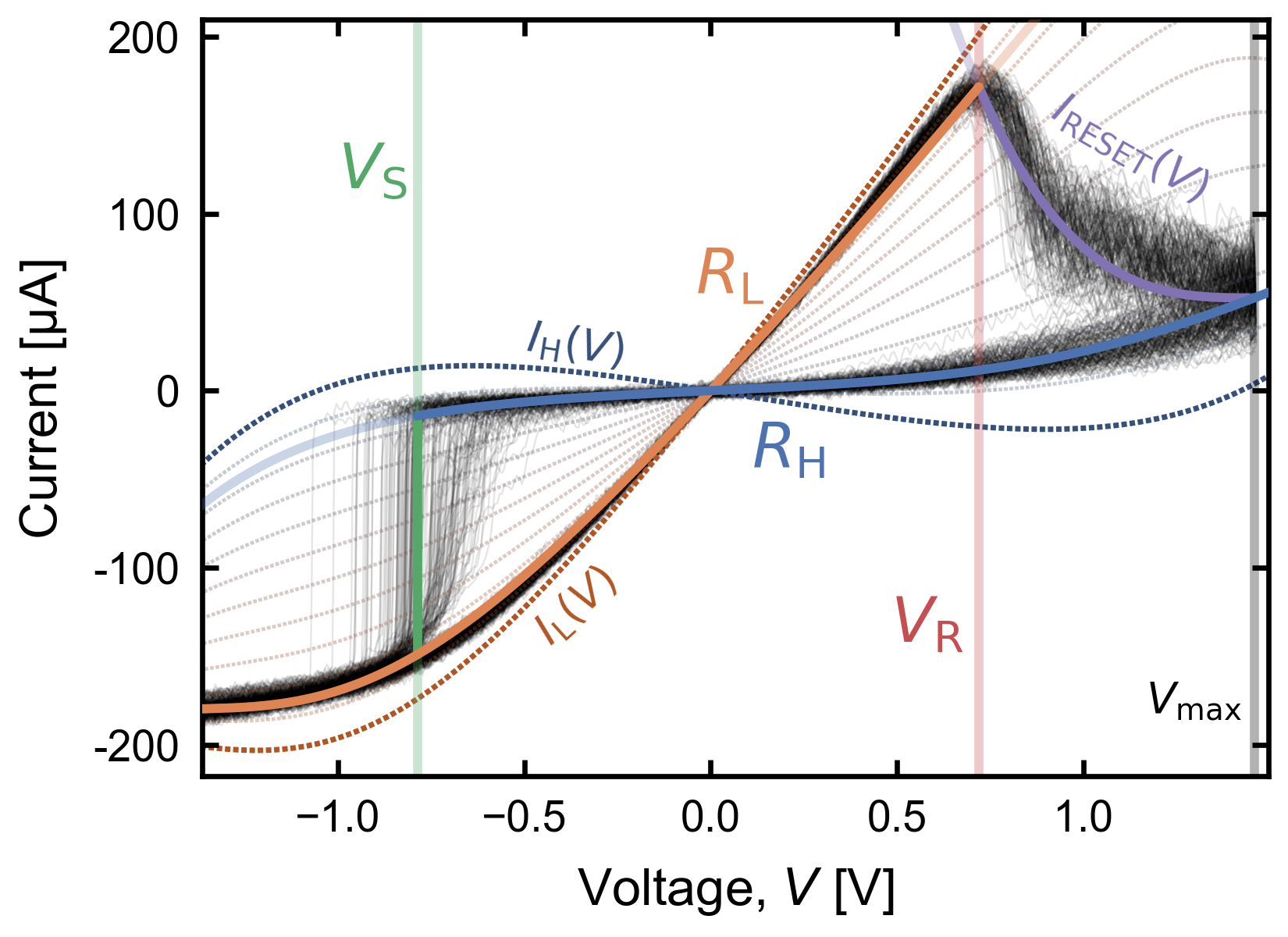}
\caption{
An exemplary $I,V$ cycle reconstructed from its feature vector representation.
States are bounded by polynomials $\IH(V)$ and $\IL(V)$. Intermediate states
(weights) are programmed by applying a voltage between $\UR$ and $\Umax$.
Experimental traces (black) are plotted in the background for reference.
}\label{fig:loop_reconstruction}
\end{figure}

Transitions of the state variables occur when the voltage applied to a device
exceeds the threshold levels for SET or RESET in its current cycle.
The transitions
connect each generated resistance state to the following one,
as illustrated in Fig.~\ref{fig:loop_reconstruction}.
Below the SET threshold $\USn$, there is an instantaneous transition from resistance state $\RHn$ to $\RLn$.
After SET has occured,
voltages above $\URn$
gradually shift the resistance state from $\RLn$ to $\RHnp$.
This gradual RESET proceeds such that
the device current
has an empirical functional form
\begin{equation}
I_{\mathrm{RESET}}(V) = a\left(\Umax - V\right)^\eta + c
\end{equation}
where 
\begin{align}
a &= \frac{I_{\mathrm{LRS},n}(\URn) - I_{\mathrm{HRS},n+1}(\Umax)}{\left(\Umax - \URn\right)^\eta}, \\
c &= I_{\mathrm{HRS},n+1}(\Umax),
\end{align}
$\Umax$ is the maximum voltage applied in the experimental sweeps,
and the constant $\eta \approx 3$ sets the curvature of the RESET transition
as estimated by a least squares fit to the training data.

\subsection{Implementation and benchmarks}

By using easily evaluated
polynomials and matrix multiplications throughout,
Synaptogen is
designed for high throughput and parallelization. 
We recently benchmarked an implementation in the Julia programming language,
comparable with the present model in terms of speed, demonstrating the
practicality of simulating large-scale physical neural networks with over $10^9$
weights~\cite{b1}.
However, due to the growing interest in simulating networks
at the circuit level, efficient stochastic device models implemented in a hardware description language (HDL)
are currently highly sought after~\cite{b4,b5,b7,b8}.

To suit a circuit design ecosystem and to compare speeds with alternative
models, we implemented Synaptogen in the \hbox{Verilog-A} HDL.\@
Special programming requirements were imposed by 
the
adaptation to a transient model description
and by the weak support for dynamic structures in \hbox{Verilog-A}.
Furthermore, due to the discontinuities at the threshold voltages, the
simulation step size was limited locally at each device threshold to aid convergence.
The order of the VAR process in the Verilog-A implementation was fixed to $p=10$.

Simulation speeds were compared with a minimalistic non-stochastic linear ion drift model (LinearDrift) as a
baseline~\cite{b3}, as well as the more complex physics-based JART v1b variability-aware
model~\cite{b4}.
Read
speeds are also compared with randomly initialized arrays of
ohmic resistances, a linearly solvable problem which gives an upper bound for the speed of the simulation framework.

We benchmarked the read and write performance for
both parallel operation of $M$ independent cells as well as for
\hbox{$\sqrt{M} \times \sqrt{M}$} crossbar arrays with resistive leads
(5~$\unit\ohm$ between every circuit node).
This distinction is important because lead resistance
has a strong impact on the system,
but is much slower to solve due to the strongly coupled equations~\cite{b10,b11}.\@
For the purpose of comparing simulation speeds between the independent-device and crossbar-connected cases,
the same applied 
voltage waveform was shared by rows and columns of the independent devices
as though they were connected by WLs and BLs.
This makes the problem equivalent to a crossbar array with zero lead resistance,
but in practice is significantly faster than enforcing crossbar connectivity in the netlist.

\begin{figure}[t!]
\centering
\includegraphics[scale=1]{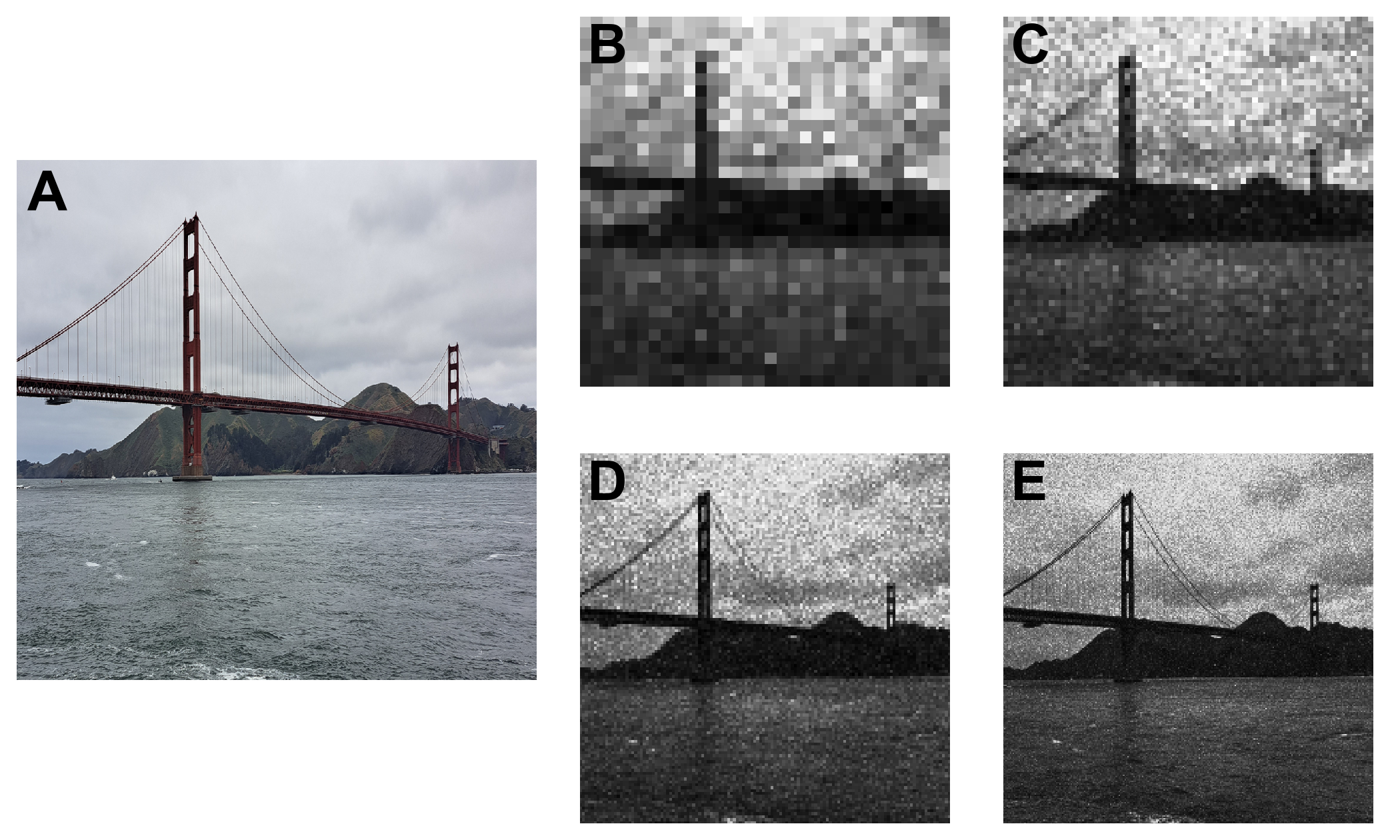}
\caption{
For write tests, the image \A was desaturated, resampled, and written into square crossbar arrays (0~\unit{\ohm} lead resistance) of
different sizes using the Verilog-A implementation of Synaptogen.
The array dimensions shown are \B 32$\times$32, \C 64$\times$64, \D 128$\times$128,
and \E 256$\times$256.
}\label{fig:image_writing_reading}
\end{figure}

\begin{figure}[t!]
\centering
\includegraphics[]{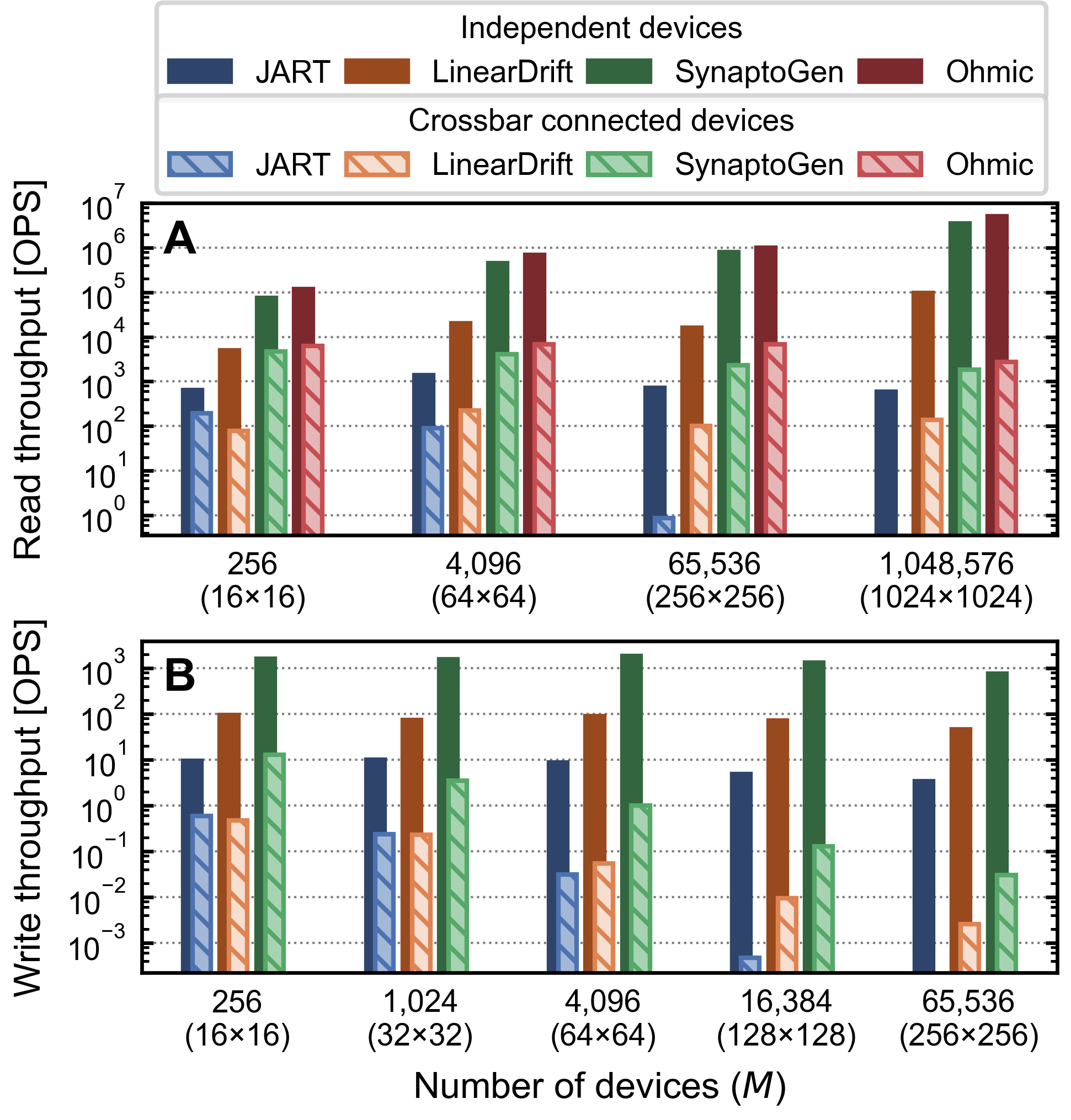}
\caption{
Benchmarks of different Verilog-A models for \A reading and \B writing
$M$ independent devices and $\sqrt{M} \times \sqrt{M}$ resistive crossbars.
For the largest arrays, the JART model did not terminate.
}\label{fig:HDL_crossbar_benchmark}
\end{figure}

Simulations were performed using the Cadence Spectre simulator with ``moderate'' settings for both
the ``accelerated parallel simulator'' (APS) and error tolerance,
running on 8 (out of 18) cores of Intel Xeon Gold 6154 CPU.\@
Square bipolar voltage pulses were applied to the WL terminals to simulate read/write operations, and the
throughput in operations per second (OPS) was calculated as the number of
devices involved in the read/write process divided by the total time taken for the transient analysis.

\begin{figure*}[ht!]
  \centering
  \includegraphics[]{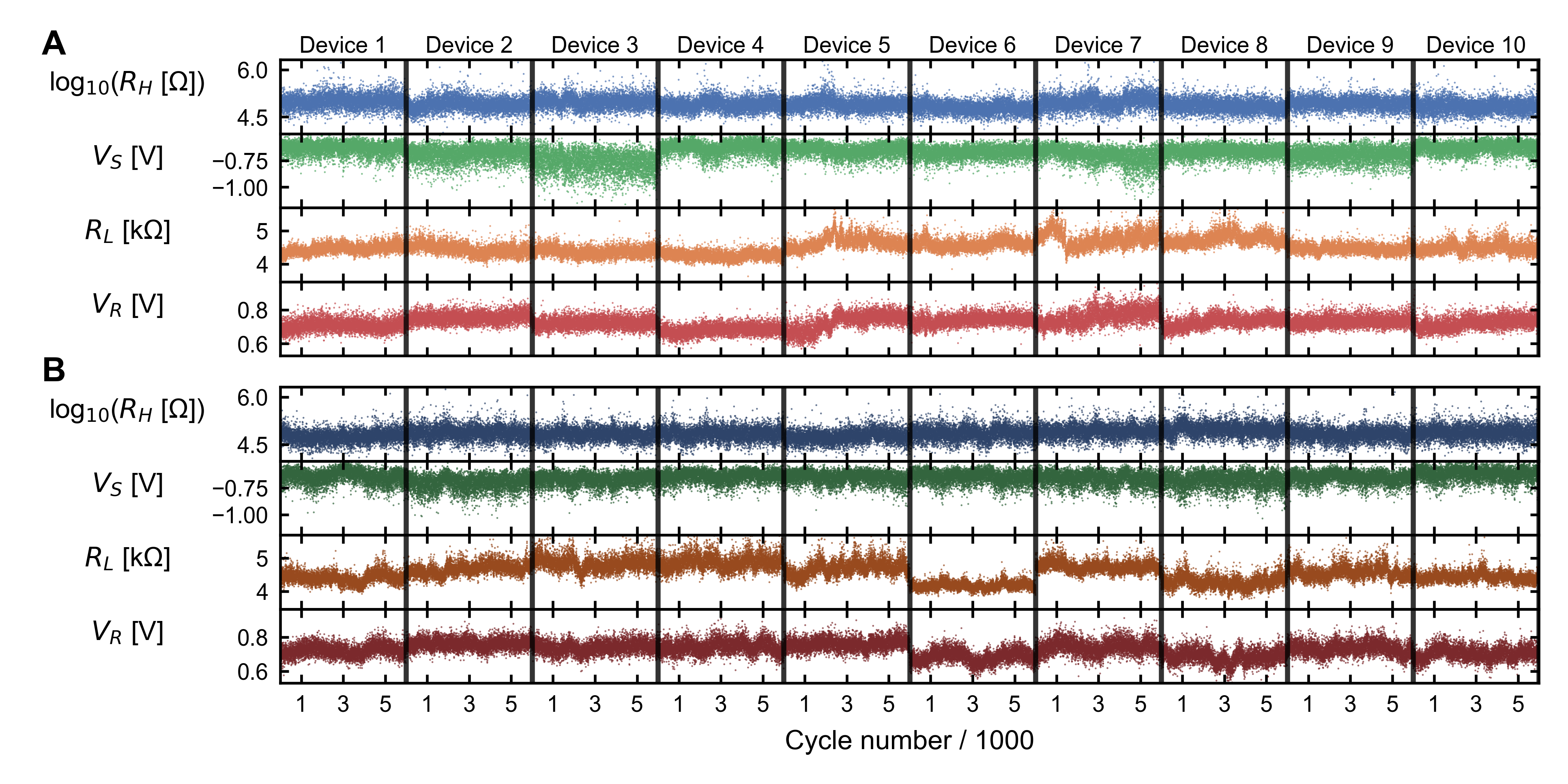}
   \caption{
A comparison between measured \A and generated \B feature vector time series across 6,000
cycles for 10 randomly selected devices.
Visual inspection confirms that the variability between devices 
and the cycling cross-correlations are closely reproduced by the model.
    }\label{fig:feature_scatter}
\end{figure*}

For weight programming benchmarks, arrays of devices were initialized in their LRS 
before writing grayscale image data into their resistance states by partial RESET (Fig.~\ref{fig:image_writing_reading}).
The pixel values were linearly mapped to a suitable reset voltage range
and, using a half-select voltage scheme~\cite{b12}, the voltages were sequentially applied to the corresponding cells for 1~\unit{\micro\second}.
For situations
where
the entire array could not be written in a practical amount of time, 
the throughput was determined by writing a 16×16 sub-block of devices.

For readout benchmarks, 200~mV pulses were simultaneously applied to all WLs for
1~ns as current was measured at the grounded BL terminals.
The results of the read and write benchmarks are summarized in Fig.~\ref{fig:HDL_crossbar_benchmark}.

\section{Results}

\begin{figure}[ht!]
  \centering
  \includegraphics[width=0.96\linewidth]{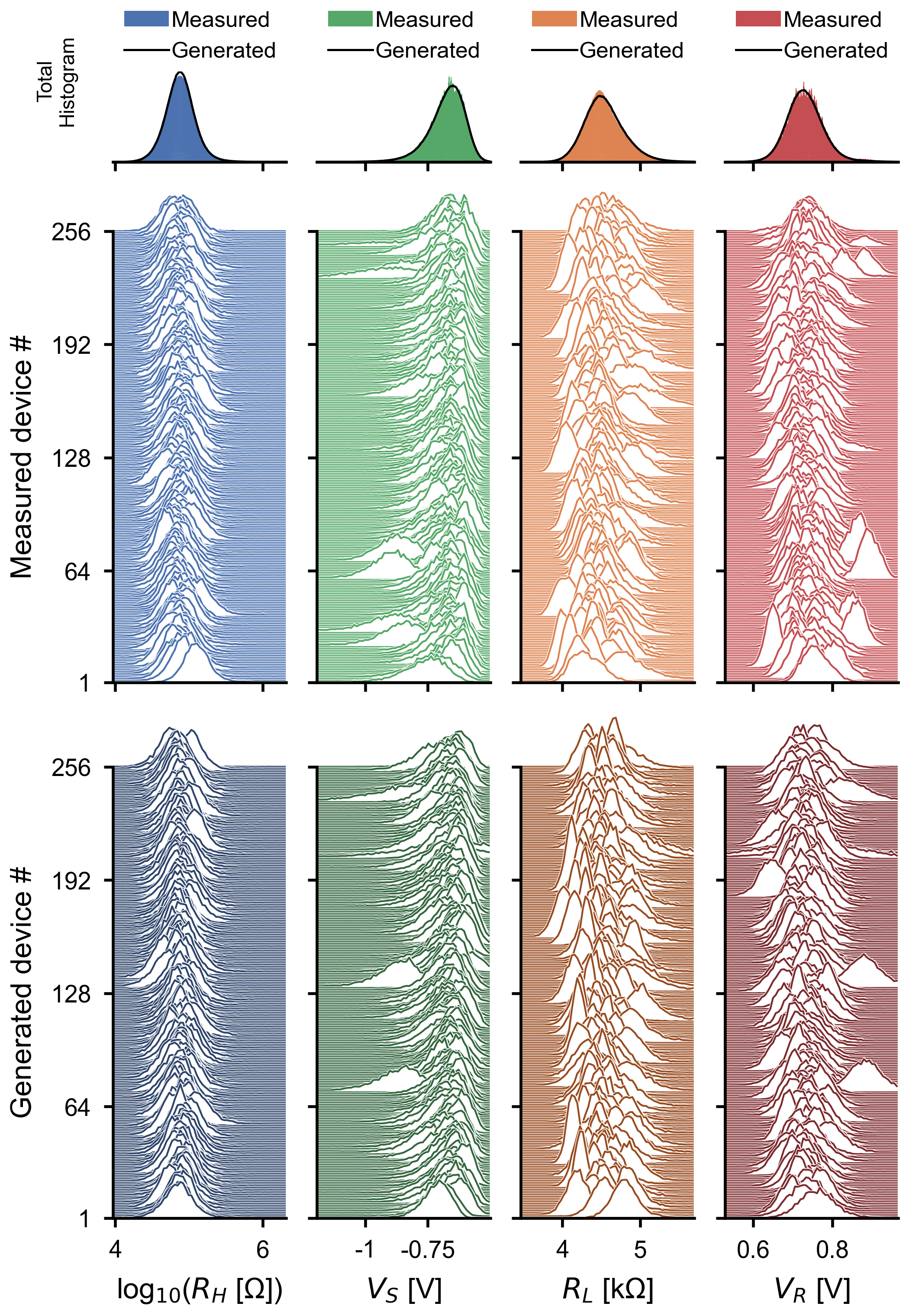}
  \caption{
A comparison of
measured and generated feature distributions for 6,000 cycles of 256
devices.
The marginal densities, their variation between
devices, as well as the total distribution across all cycles of all devices are
very closely replicated by the generative model.}\label{fig:dist_of_dists}
\end{figure}

The described hierarchical modeling approach efficiently generates feature vectors that closely resemble the training data.
This can be visually
verified with respect to the
time series behavior
(Fig.~\ref{fig:feature_scatter}).
The correlated variations in the feature distributions across different devices are very closely replicated while
simultaneously recreating the total distributions over all devices and cycles (Fig.~\ref{fig:dist_of_dists}).

For all models and conditions,
read operations were significantly faster than writes, and speeds were much higher for independent devices than for an equal number of crossbar connected devices.
Synaptogen wrote at $\sim 10^3$ OPS for independent devices,
but started at 13~OPS for 16$\times$16 crossbars, degrading with crossbar size to only 0.3 OPS at 256$\times$256.
For readout, Synaptogen is competitive with simple ohmic resistive networks, reaching 60\% to 80\% of their speed in most cases.
The throughput of these read operations increased for larger numbers of devices,
with $8 \times 10^3$~OPS for 256 devices and $4 \times 10^6$~OPS for 1,048,576 devices.
Crossbar connected readouts were slowed by
2 to 4 orders of magnitude relative to independent devices
as the array size increased from 16$\times$16 to 256$\times$256.

Synaptogen was between 10$\times$ and 100$\times$ faster than LinearDrift for all benchmarks,
which is remarkable because LinearDrift is a very simple ordinary differential equation (ODE) formulation for which the simulator should be well adapted.
Furthermore, LinearDrift does not include C2C or D2D variability, and cannot reproduce many important switching features of actual devices.
Synaptogen even more significantly outperforms the JART v1b variability model, which is more closely comparable in terms of covered device behavior.
Due to its complexity and implicit formulation, JART performance degrades faster than the other models as the array size grows;
for JART array sizes 256$\times$256
and above,
not even a single write operation could be performed in a reasonable time frame.
At 1024$\times$1024, read operations were also impossible.
For the conditions that could be simulated, 
operations on independent Synaptogen devices were always over 100$\times$ faster,
with the speed of writes approximately 200$\times$, and reads reaching 6,000$\times$ those of JART.\@
For the resistive crossbar simulations, Synaptogen was between 10$\times$ to 100$\times$ faster for 64$\times$64 and smaller arrays,
and between 100$\times$ and 10,000$\times$ for larger arrays.

Analog circuit simulations face
intrinsic
speed limitations due to the computation necessary at each time step to converge on
solutions to large systems of non-linear differential equations.
Even with dramatic speed increases over competing models,
simulation in Cadence Spectre with Synaptogen synapses 
is
practical for training and inference of fully connected neural network layers
only within
limits.
Table~\ref{tab:times} shows the time necessary to write a pre-trained model
and to perform an inference operation according to our benchmarks.
While many operations can be
completed
in well under a second, 
others 
(such as writing to large resistive crossbars) 
can take a considerable amount of time (hours or days).

As modern ML networks commonly exceed millions of weights, these benchmarks
highlight
the need
to extend the device model's
applicability to larger scales.
Therefore,
while the Verilog-A implementation
provides compatibility with circuit design tools,
we also implemented Synaptogen in the 
Julia programming language.
The internal operation is the same for both models, while the latter 
achieves orders of magnitude higher speed by avoiding transient calculations.

\begin{table}[t]
\centering
\caption{Estimated time required for neural network operations using Synaptogen weights in the Cadence Spectre simulator}\label{tab:times}
\begin{tabular}{cllll}
\toprule
Layer size & \multicolumn{2}{c}{Weight Initialization} & \multicolumn{2}{c}{Inference} \\
\cmidrule(lr){2-3} \cmidrule(lr){4-5}
{} &  Independent & Crossbar & Independent & Crossbar \\
\midrule
16×16     &      160 ms &     20 s &      3.4 ms &    54 ms \\
32×32     &      660 ms &  4.9 min &             &          \\
64×64     &       2.3 s &    1.1 h &      9.2 ms &    1.0 s \\
128×128   &        13 s &    1.5 d &             &          \\
256×256   &     1.4 min &     25 d &       82 ms &     28 s \\
1024×1024 &             &          &      300 ms &  9.6 min \\
\bottomrule
\end{tabular}
\end{table}

\section{Conclusion}

In this work, we developed a generative compact model for resistive switching devices that
seamlessly adapts to statistical measurements.
Through an automated training procedure, the model closely captures both C2C and D2D variability
of data measured on integrated ReRAM devices.
While an equivalent model can be used in a high-level programming domain for
larger scale simulations, here we demonstrate its use in analog circuit simulation of
1T1R arrays.
The implemented circuit level model operates orders of magnitude faster for reading and writing compared to other compact models,
and we demonstrate crossbar programming (256$\times$256 devices) and readout (1024$\times$1024 devices)
at scales which far exceed what was previously possible in the analog circuit simulation domain.

\section*{Code availability}


The Verilog-A compact model as well as its Julia counterpart are
available on GitHub (\url{https://github.com/thennen/synaptogen}) and
archived in Zenodo (\url{https://zenodo.org/doi/10.5281/zenodo.10942560}).


\begin{thebibliography}{00}
\bibitem{b0} 
C. Nail et al., “Understanding RRAM endurance, retention and window margin trade-off using experimental results and simulations,” in 2016 IEEE International Electron Devices Meeting (IEDM), San Francisco, CA, USA: IEEE, Dec.\ 2016, p. 4.5.1-4.5.4.\ doi: 10.1109/IEDM.2016.7838346.
\bibitem{b1} 
T. Hennen et al., “A high throughput generative vector autoregression model for stochastic synapses,” Front. Neurosci., vol. 16, p. 941753, Aug. 2022, doi: 10.3389/fnins.2022.941753.
\bibitem{b2} 
A. Grossi et al., “Fundamental variability limits of filament-based RRAM,” in 2016 IEEE International Electron Devices Meeting (IEDM), San Francisco, CA, USA: IEEE, Dec. 2016, p. 4.7.1-4.7.4. doi: 10.1109/IEDM.2016.7838348.
\bibitem{b6} 
J. D. Hamilton, Time series analysis. Princeton, N.J: Princeton University Press, 1994.
\bibitem{b4} 
C. Bengel et al., “Variability-Aware Modeling of Filamentary Oxide-Based Bipolar Resistive Switching Cells Using SPICE Level Compact Models,” IEEE Transactions on Circuits and Systems I: Regular Papers, vol. 67, no. 12, pp. 4618–4630, 2020, doi: 10.1109/TCSI.2020.3018502.
\bibitem{b5} 
V. Ntinas et al., “A Simplified Variability-Aware VCM Memristor Model for Efficient Circuit Simulation,” in 2023 19th International Conference on Synthesis, Modeling, Analysis and Simulation Methods and Applications to Circuit Design (SMACD), Funchal, Portugal: IEEE, Jul. 2023, pp. 1–4. doi: 10.1109/SMACD58065.2023.10192107.
\bibitem{b7}
J. Reuben, M. Biglari, and D. Fey, “Incorporating Variability of Resistive RAM in Circuit Simulations Using the Stanford–PKU Model,” IEEE Trans. Nanotechnology, vol. 19, pp. 508–518, 2020, doi: 10.1109/TNANO.2020.3004666.
\bibitem{b8}
S. Guitarra, P. Mahato, D. Deleruyelle, L. Raymond, and L. Trojman, “Stochastic based compact model to predict highly variable electrical characteristics of organic CBRAM devices,” Solid-State Electronics, vol. 185, p. 108055, Nov. 2021, doi: 10.1016/j.sse.2021.108055.
\bibitem{b3} 
S. Kvatinsky, E. G. Friedman, A. Kolodny, and U. C. Weiser, “TEAM: ThrEshold Adaptive Memristor Model,” IEEE Trans. Circuits Syst. I, vol. 60, no. 1, pp. 211–221, Jan. 2013, doi: 10.1109/TCSI.2012.2215714.
\bibitem{b10}
A. Chen, “A Highly Efficient and Scalable Model for Crossbar Arrays with Nonlinear Selectors,” in 2018 IEEE International Electron Devices Meeting (IEDM), San Francisco, CA: IEEE, Dec. 2018, p. 37.2.1-37.2.4. doi: 10.1109/IEDM.2018.8614505
\bibitem{b11}
D. Joksas and A. Mehonic, “badcrossbar: A Python tool for computing and plotting currents and voltages in passive crossbar arrays,” SoftwareX, vol. 12, p. 100617, Jul. 2020, doi: 10.1016/j.softx.2020.100617.
\bibitem{b12}
A. Chen, “Analysis of Partial Bias Schemes for the Writing of Crossbar Memory Arrays,” IEEE Trans. Electron Devices, vol. 62, no. 9, pp. 2845–2849, Sep. 2015, doi: 10.1109/TED.2015.2448592.
\end{thebibliography}
\end{document}